\documentclass[letterpaper]{article} 
\usepackage{aaai24}  
\usepackage{times}  
\usepackage{helvet}  
\usepackage{courier}  
\usepackage[hyphens]{url}  
\usepackage{graphicx} 
\usepackage{natbib}  
\usepackage{caption} 
\usepackage{algorithm}
\usepackage{algorithmic}

\usepackage{newfloat}
\usepackage{listings}
\DeclareCaptionStyle{ruled}{labelfont=normalfont,labelsep=colon,strut=off} 
\floatstyle{ruled}
\newfloat{listing}{tb}{lst}{}
\floatname{listing}{Listing}
\usepackage{array,multirow}
\usepackage{booktabs}
\usepackage{amsmath}

\title{Teaching Large Language Models to Translate with Comparison}
\author{
    Jiali Zeng, Fandong Meng,
    Yongjing Yin, Jie Zhou
}
\affiliations{
    Pattern Recognition Center, WeChat AI, Tencent Inc\\
    \{lemonzeng,fandongmeng,yongjingyin,withtomzhou\}@tencent.com
}

\usepackage{bibentry}

\begin{document}

\maketitle

\begin{abstract}

Open-sourced large language models (LLMs) have demonstrated remarkable efficacy in various tasks with instruction tuning. 
However, these models can sometimes struggle with tasks that require more specialized knowledge such as translation. 
One possible reason for such deficiency is that instruction tuning aims to generate fluent and coherent text that continues from a given instruction without being constrained by any task-specific requirements. 
Moreover, it can be more challenging to tune smaller LLMs with lower-quality training data.
To address this issue, we propose a novel framework using examples in comparison to teach LLMs to learn translation. 
Our approach involves output comparison and preference comparison, presenting the model with 
carefully designed examples of correct and incorrect translations and an additional preference loss for better regularization.
Empirical evaluation on four language directions of WMT2022 and FLORES-200 benchmarks shows the superiority of our proposed method over existing methods. 
Our findings offer a new perspective on fine-tuning LLMs for translation tasks and provide a promising solution for generating high-quality translations.
Please refer to Github for more details:
https://github.com/lemon0830/TIM.

\end{abstract}

\section{Introduction}

Generative large language models, like GPT models, have shown impressive performance in various NLP tasks \cite{gpt3,conf/nips/Ouyang0JAWMZASR22}, including machine translation \cite{Hendy,Zhu},
which opens up new possibilities for building more effective translation systems.
It is impractical to deploy such large models for the translation task only, and using or tuning open-sourced generative language models has become an attractive research direction.
In this regard, researchers have explored strategies for example selection and instruction design through In-Context Learning (ICL) \cite{emnlp/LinMAWCSOGBDPSK22,Agrawal}. 
However, evaluations of open-sourced LLMs show that they do not perform as well as strong multilingual supervised baselines in most translation directions \cite{Zhu}. 
Additionally, ICL can increase decoding latency due to longer context. 
Based on these observations, researchers suggest tuning relatively small LLMs for translation with a few high-quality supervised instructions \cite{Hendy,zeng2023improving,Jiao_ParroT}.

Instruction tuning is an efficient method for making LLMs better aligned to the task descriptions preferred by humans \cite{conf/nips/StiennonO0ZLVRA20,conf/nips/Ouyang0JAWMZASR22,flant5,selfinstruct}.
The only requirement is to collect task-specific data, on which LLMs will be fine-tuned with the language modeling loss.
However, optimizing for simple next-token prediction loss will cause models to overlook context information, especially for low-capacity models.
It is serious for the tasks in which the specialized knowledge in context is necessary for task completion (e.g., translation), and ignoring such knowledge on translation can lead to inadequacy and hallucination.
Therefore, there is a need to investigate the limitations of LLMs and explore methods for improving their performance in specialized tasks.

In this paper, we propose to {\bf T}each the language models to learn translation w{\bf I}th examples in co{\bf M}parison, named {\bf TIM}, aiming to make full use of a small amount of high-quality translation data.
Based on the training data, we further construct two kinds of comparisons: 
output comparison and preference comparison.
Output comparison is used to learn responses to different instructions for the same input.
Preference comparison is used to maximize the gap between correct and incorrect translations.
Specifically, to help identify specific areas where the model may be making errors, we introduce an additional preference loss, 
which is originally used to learn reward models \cite{conf/nips/StiennonO0ZLVRA20}, as regularization to penalize unexpected outputs.

We evaluate our proposed method on WMT22 and FLORES-200 test sets (EN$\Leftrightarrow$DE, EN$\Leftrightarrow$ZH), and the improvement over the baselines shows the effectiveness of our method.
Our model shows better zero-shot translation performance and stability in prompt choice. 
Moreover, the performance of the models tuned by our TIM increases as the model size increases, with the improvement being more pronounced in the case of smaller models.
In particular, the tuned LLaMA-2-13B \cite{arxiv2023:LLaMA} achieves top 1 on quality estimation without references in the EN$\Leftrightarrow$DE, outperforming the dedicated models for quality estimation. 


\begin{figure*}[!th]
\centering
\includegraphics[width=1.0\linewidth]{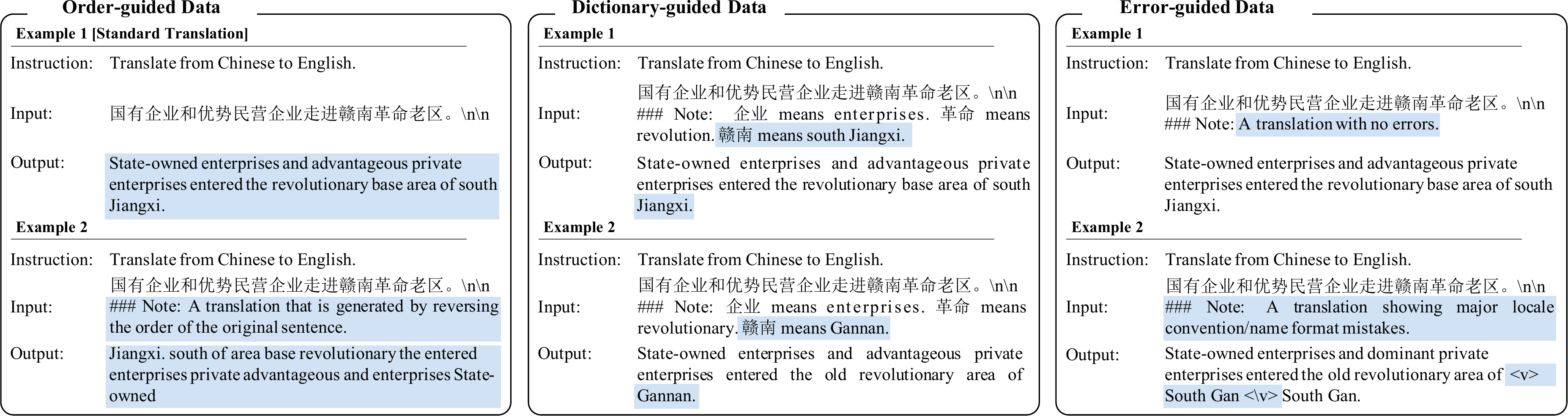}
\caption{
{Illustration of three types of output comparison.}
The text in blue highlights the difference between the added notes and the resulting difference due to these specific notes.
}
\label{fig_data_comparison}
\end{figure*}

\section{Method}
In brief, we tune generative language models to learn translation with output comparison and preference comparison in the instruction tuning framework.
First, we will give a formal introduction to instruction tuning.
Then, we present the details of two kinds of comparisons of our method consisting of output comparison and preference comparison and an additional preference learning loss.
Finally, we explore different approaches to parameter tuning.

\subsection{Background: \ Instruction Tuning}
Instruction tuning aims to enhance the capacity of language models to handle instructions in natural languages.
The concept is that the models can be trained to execute tasks specified in instructions, which would enable them to comprehend the tasks and even process tasks not encountered before.

Generally, each instance of instruction-following data starts with ``instructions'' $c$ describing a task, and a corresponding output $y$ indicating the answer to the instruction.
The ``input'' $x$, the optional context or input for the task, is not necessary but is required for the machine translation task.
Given the instruction data, the language models are optimized by minimizing the negative log-likelihood of the output $y$:
\begin{equation}
    L_{lm}=-\frac{1}{|y|}\sum_i^{|y|}\text{logp}(y_i|c,x).
\end{equation}
Notably, the objective is the same as that used in pretraining.

\subsection{Output Comparison}
\label{sec_output_comparison}
An important ingredient of our method is the construction of samples used to provide comparison signals for model learning.
In addition to regular translation data, we construct data used for comparison by introducing sequence ordering, dictionary information, or translation errors, which are shown in Figure \ref{fig_data_comparison}. 

\begin{figure*}[!th]
\centering
\includegraphics[width=0.9\linewidth]{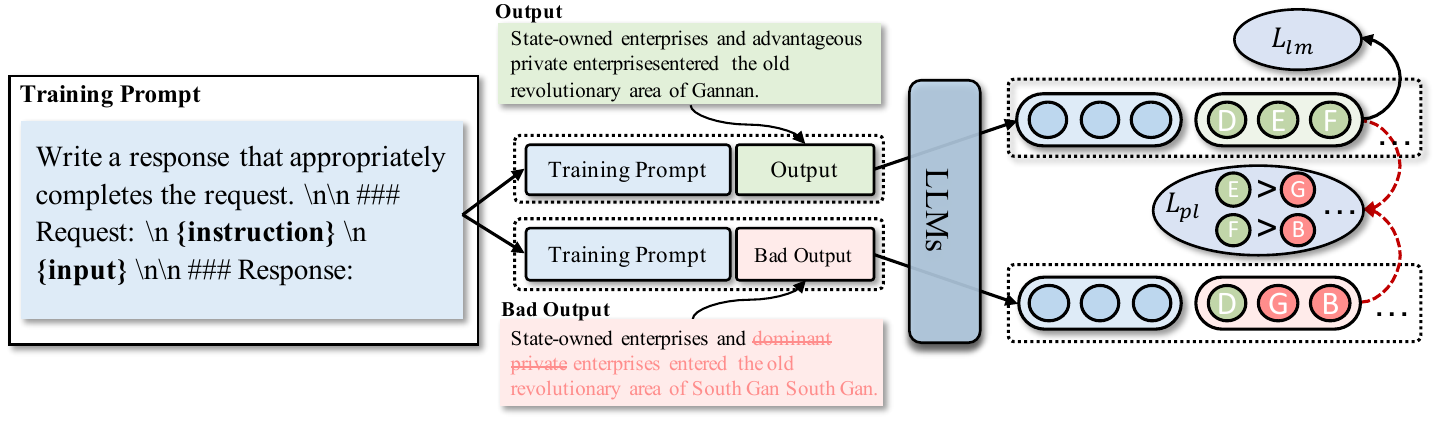}
\caption{
{Overall framework of our proposed TIM.}
Given the contrastive outputs of each instance, we optimize the LLMs with the general language modeling loss and
the token-level preference loss.}
\label{fit_model}
\end{figure*}

\begin{figure}[!t]
\centering
\includegraphics[width=0.95\linewidth]{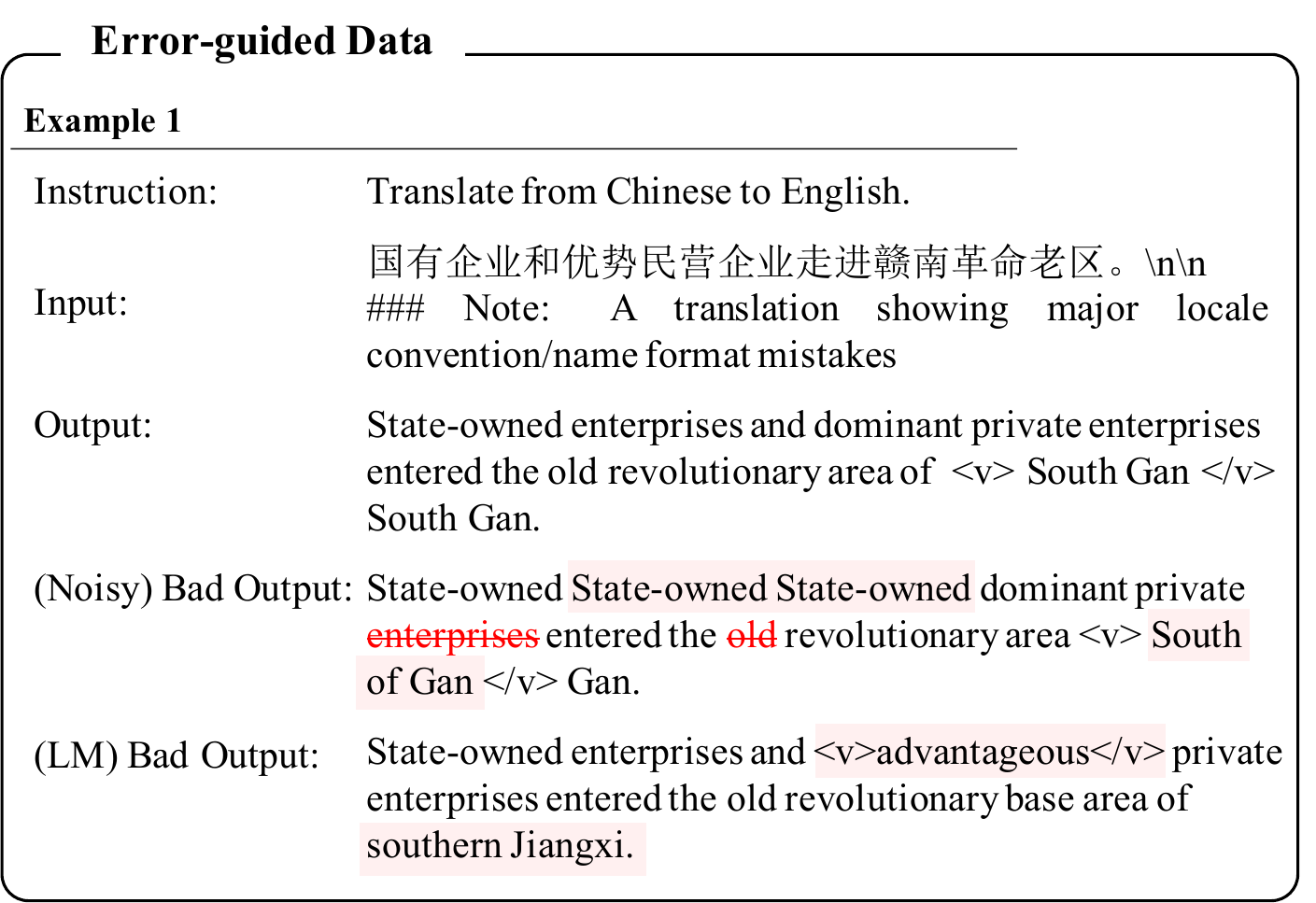}
\caption{{An example of contrastive outputs for preference Comparison.}
The ``Bad Output'' denotes the noisy translation used to be compared with the ``Output''.}
\label{fig_preference_comp}
\end{figure}

\paragraph{Order-guided data.}
We introduce a variation of the translation process, and
we reverse the translations of certain examples and provide an accompanying note indicating the reverse generation order (Order-guided Data in Figure \ref{fig_data_comparison}).
By training on these reverse sentences, the model gains the ability to capture dependencies that may not be evident in the original sentence order.
This helps improve the model's comprehension of instructions and enhances its capability to generate coherent and contextually appropriate translations.

\paragraph{Dictionary-guided Data.}
To make the model aware of the underlying reasons for different translations, we inform the model of different correct outputs with the help of bilingual dictionaries\footnote{https://github.com/facebookresearch/MUSE}.
Instead of synthesizing the comparison data, we utilize an existing multi-reference corpus.
By looking up the bilingual dictionary, we establish word alignments between a single source sentence and multiple references.
The word alignments serve as annotations appended to the input.
Illustrated in Figure \ref{fig_data_comparison}, 
the notes contain distinct word alignments, and the outputs of {\bf Example 1} and {\bf Example 2} differ despite the same input sentences.

\paragraph{Error-guided Data.}
We introduce translations with error annotations inspired by \citet{Jiao_ParroT}.
The added notes indicate no mistakes in the references for correct input-output pairs, while the notes of incorrect input-output pairs indicate detailed translation errors.
As shown in the right part of Figure \ref{fig_data_comparison}, 
the translation of {\bf Example 1} is correct while the translation of {\bf Example 2} 
has a major locale convention format mistake, corresponding to the added note.

\subsection{Preference Comparison}
In preference comparison, we assign contrastive outputs for each data type, denoted as {\it Bad Output}, and train the model with an extra preference loss.
As illustrated in Figure \ref{fig_preference_comp}, we propose two types of the {\it Bad Output}: 1) {\bf Noisy-based}, in which we intentionally introduce noise into the original output by randomly deleting words or swapping the positions of two words; 
2) {\bf LM-based}, in which we fine-tune a relatively small LM (e.g., BLOOM-1b7) and generate output using a simple sampling strategy for each instance.
With examples of correct and incorrect translations, the model is optimized to distinguish higher-quality translations, which can reduce the resource requirement for training.

One way to utilize the contrastive outputs is to train a reward model and further fine-tune language models with the reward model using reinforcement learning, i.e., RLHF \cite{conf/nips/StiennonO0ZLVRA20,conf/nips/Ouyang0JAWMZASR22}.
Instead of using such a complex two-stage training process, we directly tune the language model using a token-level preference loss:
\begin{equation}
    L_{pl}=-\frac{1}{N-I}\sum_{i=I}^Nmax(0,-r_{\theta}(h_i^{(0)})+r_{\theta}(h_i^{(1)})+1.0),
\end{equation}
where 
$N$ is the maximum length of two sequences, and $h^{(0)}_i$ and $h^{(1)}_i$ are the hidden state of the $i$-th token of the preferred output $y_0$ and comparison output $y_1$, respectively.
$I$ is the index starting from the segments different between $y_0$ and $y_1$.
As illustrated in Figure \ref{fig_preference_comp}, the overlapping part of Output and Bad Output ``State-owned enterprises and'' will not contribute to the calculation of $L_{pl}$.
Specifically, $r_{\theta}$ is a linear head that takes the hidden state of the top layer and returns a scalar.


The overall loss function for tuning the model is 
\begin{equation}
    L=L_{lm}+{\lambda}L_{pl},
\end{equation}
where $\lambda$ is a coefficient of the preference learning loss. 
We simply set $\lambda$ as 1.0 in this paper.

\subsection{Tuning Strategies}
In this paper, we adopt three different strategies for fine-tuning, listed in descending order from the number of trainable parameters.

\paragraph{LoRA: Tuning with Low-rank Matrices.}
LoRA \cite{hu-etal-2022-LoRA} is a technique that reduces the number of trainable parameters by introducing new low-rank matrices to any module in the model while keeping the original weights frozen. 
This results in a significant reduction in storage requirements and efficient task-switching during deployment without impacting inference latency.

\paragraph{FixEmb: Tuning with Embedding Fixed.}
LoRA-based tuning has a limitation where the limited number of trainable parameters may restrict its expressiveness.
A simple solution to overcome this is to fine-tune the parameters of the model layers while keeping the embeddings fixed. 
This allows the model to gain more flexibility in adjusting its performance without compromising the semantic information captured by the embeddings. 

\paragraph{Full: Tuning Full Parameters.}
Full parameter tuning has recently been demonstrated more effective than LORA.
The major limitation of full parameter fine-tuning is the memory footprint, but it is not serious for 7B models and little data.

\section{Experiments}

In this section, we begin by conducting preliminary experiments to investigate the impact of inference strategies and the resilience of our TIM under varying instructions. 
Subsequently, we evaluate TIM on the WMT and FLORES-200 dev-test tasks
in four language directions. 

\subsection{Settings}

To avoid data leakage \cite{arxiv2023:unreasonable}, we use the latest WMT22 test set and FLORES-200 dev-test:
    1) 
    We use the test sets from WMT22 competition\footnote{https://www.statmt.org/wmt22/translation-task.html}, 
    which consist of more recent content from diverse domains such as news, social, e-commerce, and conversational domains.
    The test sets comprise 1984, 2037, 1875, and 2037 samples for the German-to-English (De$\Rightarrow$En), English-to-German (En$\Rightarrow$De), Chinese-to-English (Zh$\Rightarrow$En), and English-to-Chinese (En$\Rightarrow$Zh) language pairs, respectively.
    2) 
    We use the dev-test split from the FLORES-200 benchmarks\footnote{https://github.com/facebookresearch/flores/blob/main/flores200}.
    This dataset includes 1,012 sentences extracted from English Wikipedia, covering a broad range of topics and domains. 
    Professional translators have carefully checked these sentences into approximately 200 languages.

To ensure a fair and consistent evaluation, we fine-tuned all models for 1 epoch with a batch size of 128, while imposing a maximum text length of 512. 
The learning rates are 2e-5 for FixEmb and Full, and 3e-4 for LoRA, respectively.
The weight decay parameter is set to 0.0.
We conducted fine-tuning on eight NVIDIA A100 GPUs, utilizing the Deep-Speed ZeRO stage3 for model parallelism. 
The results of the final checkpoints are reported.
For automatic evaluations, we utilize two widely adopted metrics: BLEU \cite{papineni2002bleu:} implemented in SacreBLEU\footnote{https://github.com/mjpost/sacrebleu}, and COMET\footnote{https://github.com/Unbabel/COMET} with {\it Unbabel/wmt22-comet-da}. 
BLEU is driven by n-gram similarity, while COMET relies on cross-lingual pre-trained models.

\subsection{Baselines}

We leverage {\bf BLOOMZ-7b-mt}\footnote{https://huggingface.co/bigscience/bloomz-7b1-mt} and {\bf LLaMA-2-7b}\footnote{https://huggingface.co/meta-llama/Llama-2-7b} \cite{touvron-etal-2023-llama2} as the backbones and evaluate the following baselines:

\paragraph{Alpaca-(*)} is a reproduction of the Alpaca model fine-tuned solely on the alpaca multi-task dataset\footnote{https://huggingface.co/datasets/tatsu-lab/alpaca}.

\paragraph{MT-(*)} is fine-tuned on the human-written validation data from previous WMT competitions, i.e., the newstest2017-2021 of Chinese$\Leftrightarrow$English and German$\Leftrightarrow$English, which consist of 45,433 sentence pairs for all four directions.

Besides, we report the results of WMT22 winners, 
and NLLB-3.3B \cite{arxiv2022:NLLB}.  
The latter is a multilingual translation model trained on a massive parallel corpus of over 200 languages\footnote{The results in \cite{Arxiv2023:bayling} are directly reported.}.
We use the notation {\bf TIM-(*)} to refer to LLMs fine-tuned using our proposed TIM approach. 
In practice, to construct the order-guided data, we utilize the WMT translation data.
Besides,
we rely on the annotated data of newstest2020 Zh$\Rightarrow$En and En$\Rightarrow$De in the Multidimensional Quality Metrics (MQM) datasets.
We use the mqm\_newstest2020\_ende.tsv and mam\_newstest2020\_zhen.tsv to construct the ``Error-guided’’ data\footnote{https://github.com/google/wmt-mqm-human-evaluation/tree/main/newstest2020}.
Specifically, we consider the column ``severity'', where treat the ``No-error'' label as the translations without error and others as the translations with errors.
The training data for TIM-(*) consists of the alpaca dataset, the WMT translation data, the Dictionary-guided data, Order-guided data constructed from the WMT validation data, and Error-guided data constructed from MQM data.

\begin{table}[!t]
\centering
\small
\setlength{\tabcolsep}{1.4mm}{
\begin{tabular}{lcccc}
\toprule
{\bf Method}
& {\bf Zh$\Rightarrow$En} & {\bf En$\Rightarrow$Zh} & {\bf De$\Rightarrow$En} & {\bf En$\Rightarrow$De} \\
\midrule
Sample & 22.75 & 34.98 & 24.72 & 19.09 \\
\ \ w/ {\it No Err.} & 23.10 & 36.37 & 25.20 & 19.34 \\
\ \ w/ {\it Dict.} & 21.28 & 34.55 & 24.37 & 18.19 \\
Beam-4 & {24.51} & {37.83} & 26.12 & {20.90} \\
\ \ w/ {\it No Err.} & 24.26 & {\bf 38.17} & {\bf 26.24} & {\bf 21.10} \\
\ \ w/ {\it Dict.} & {\bf 24.55} & 36.32 & 26.16 & 20.19 \\
\bottomrule
\end{tabular}}
\label{tab_results_inference_strategy}
\caption{
{Effect of inference strategies}.
We fine-tune BLOOMZ-7b-mt with our TIM and report BLEU scores on four language pairs.
}
\end{table}

\begin{figure}[!t]
\centering
\includegraphics[width=0.85\linewidth]{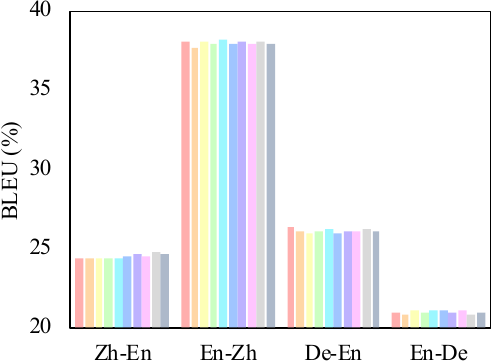}
\caption{
{\bf Effect of instructions.} We fine-tune  BLOOMZ-7b-mt with our TIM and report BLEU scores of 10 different instructions on four language pairs.}
\label{fig_diff_prompt}
\end{figure}

\subsection{Pre-Experiments}

\begin{table*}[!th]
\centering
\small
\begin{tabular}{lcccccccc}
\toprule
\multirow{2}{*}{\bf Model} &
 \multicolumn{2}{c}{\bf Zh$\Rightarrow$En} & \multicolumn{2}{c}{\bf En$\Rightarrow$Zh} & \multicolumn{2}{c}{\bf De$\Rightarrow$En} & \multicolumn{2}{c}{\bf En$\Rightarrow$De} \\
& {BLEU} & {COMET} & {BLEU} & {COMET} & {BLEU} & {COMET} & {BLEU} & {COMET} \\
\midrule
\multicolumn{9}{c}{{\bf Test:} {\it WMT22 Test Sets} \ \ \  {\bf Backbone:} {\it BLOOMZ-7b-mt}} \\
WMT22 Winners$^{*}$ & 33.5 & 81.0 & 54.3 & 86.8 & 33.7 & 85.0 & 38.4 & 87.4 \\
NLLB-3.3b$^{*}$ & 21.07 & 76.92 & 32.52 & 81.56 & 29.54 & 83.42 & 33.98 & 86.23 \\
\midrule
Alpaca-LoRA & 12.61 & 76.36 & 24.30 & 81.18 & 16.04 & 71.17 & 8.05 & 57.54 \\
Alpaca-Full & 13.01 & 75.95 & 20.65 & 78.69 & 16.98 & 72.46 & 2.28 & 36.91 \\
MT-LoRA & 21.47 & 79.20 & 35.22 & 85.00 & 23.59 & 76.91 & 15.74 & 66.42 \\
MT-FixEmb & 23.08 & 78.95 & 37.09 & 85.02 & 24.99 & 78.19 & 19.05 & 71.89 \\
MT-Full & 22.81 & 79.15 & 34.49 & 84.26 & 24.72 & 77.84 & 18.79 & 71.65 \\
\midrule
\multicolumn{9}{l}{\it w/ Noisy-based Bad Output} \\
TIM-LoRA & 22.11 & 78.89 & 35.70 & 84.90 & 23.55 & 76.70 & 16.46 & 66.80 \\
TIM-FixEmb & 24.11 & 79.70 & 37.46 & {\bf 85.29} & {\bf 26.20} & 78.79 & {\bf 20.97} & 74.63 \\
TIM-Full & 23.49 & 79.17 & 34.70 & 84.26 & 25.11 & 78.40 & 20.99 & 74.12 \\
\multicolumn{9}{l}{\it w/ LM-based Bad Output} \\
TIM-LoRA & 22.22 & 78.81 & 35.71 & 84.67 & 23.82 & 76.57 & 16.62 & 66.67 \\
TIM-FixEmb & {\bf 24.51} & {\bf 79.71} & {\bf 37.83} & 85.10 & 26.12 & {\bf 78.94} & 20.90 & {\bf 74.91} \\
TIM-Full & 23.81 & 79.33 & 35.57 & 84.75 & 25.43 & 78.19 & 20.74 & 74.24 \\
\bottomrule
\toprule
\multicolumn{9}{c}{{\bf Test: } {\it FLORES-200} \ \ \ {\bf Backbone: } {\it LLaMA-2-7b}} \\
MT-FixEmb & 26.41 & {\bf 85.88} & 33.80 & 84.88 & 42.14 & 88.92 & 32.23 & 86.16 \\
MT-Full & 26.06 & 85.81 & 33.75 & 84.92 & 41.56 & 88.77 & 31.71 & 85.93 \\
\midrule
\multicolumn{9}{l}{\it w/ Noisy-based Bad Output} \\
TIM-FixEmb & {\bf 26.47} & 85.64 & 34.84 & {\bf 85.47} & 42.24 & {\bf 88.95} & 33.01 & {\bf 86.32} \\
TIM-Full & 26.30 & 85.71 & 34.46 & 85.23 & 42.01 & 88.68 & 32.28 & 86.05 \\
\multicolumn{9}{l}{\it w/ LM-based Bad Output} \\
TIM-FixEmb & 26.13 & 85.61 & {\bf 35.15} & 85.27 & {\bf 42.91} & 88.84 & {\bf 33.32} & 86.20 \\
TIM-Full & 26.25 & {85.81} & 34.53 & 85.18 & 41.96 & 88.82 & 32.79 & 86.05 \\
\bottomrule
\end{tabular}
\label{tab_results_main_result}
\caption{
{Evaluation results of different LLMs on 4 language pairs from WMT22 test sets and Flores devsets}. Methods with * denote that we directly report the scores from the corresponding paper, and others are from our implementation. 
}
\end{table*}

Here, we investigate the effect of inference strategies and instructions.
We fine-tune the {\bf BLOOMZ-7b-mt} with our TIM and conduct evaluations on the WMT22 test sets.

\paragraph{Effect of Inference Strategies.} \label{sec_inference_strategy}
We compare the performance of sampling and beam search, and the two search algorithms are combined with the notes in our dictionary-guided and error-guided data.
Table \ref{tab_results_inference_strategy} presents the experimental results. 
First, we observe that instructing the model to generate translations without errors does not result in a significant performance gain.
We speculate that the preference loss function implicitly allows the LLMs to learn to generate error-free translations, making the additional instructions unnecessary.
Secondly, previous studies have shown that introducing alignment information from dictionaries can improve translation performance \cite{arxiv2023:chainofdict,zheng-etal-2021-non-parametric,arxiv2016:bridging}. 
Surprisingly, adding alignment notes 
harms the performance, and this may be due to that most of the words in the dictionaries we use are common words, or that the wording styles of the dictionaries differ greatly from the reference. 
How to better collect and use a dictionary for machine translation \cite{thompson-etal-2019-hablex} is left for future work.

\paragraph{Effect of Instructions.}
In human interaction scenarios, instructions provided by users may vary in styles and forms, and thus it is essential to evaluate the robustness of TIM under different instructions. 
We use ten distinct instructions and the result in Figure \ref{fig_diff_prompt} indicates that our TIM achieves consistent performance across all the tested instructions.

\subsection{Main Results}
Based on the observation in Section \ref{sec_inference_strategy}, we use a simple instruction ``Translate from \{src\} to \{tgt\}.\textbackslash{}n\{input\}'' and beam search with a beam size of 4 for all models during inference.
Table \ref{tab_results_main_result} presents the translation performance on the WMT22 test sets and FLORES-200 dev-test.

We have the following observations:
First, we observe significant performance fluctuations across different language models, training data, and language pairs for (*)-{\it LoRA} and (*)-{\it Full}. 
For example, with {\bf BLOOMZ-7b-mt} as the backbone, {\it Alpaca-LoRA} outperforms {\it Alpaca-Full} in most language pairs,
while {\it MT-LoRA} underperforms {\it MT-Full}.
We speculate that LoRA can prevent LLMs from overfitting but is limited in the number of trainable parameters. 
In contrast, the experiment result of (*)-{\it FixEmb} indicates that fine-tuning with fixed embedding parameters can better leverage the generalization of LLMs and prevent overfitting.
Second, training LLMs with comparison can further enhance the understanding of the translation task. 
Compared to {\it Alpaca}-(*), {\it MT}-(*) models, {\it TIM}-(*) exhibits notably better results on both the WMT22 test sets and FLORES-200 dev-test. 

\section{Analysis}

\begin{figure*}[!t]
\centering
\includegraphics[width=0.9\linewidth]{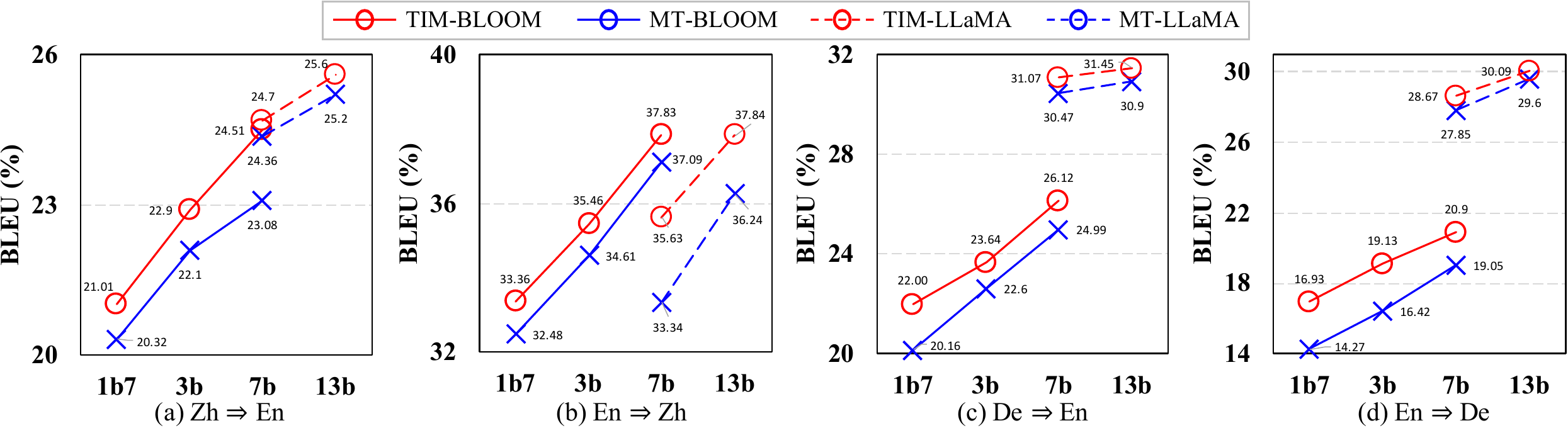}
\caption{
{Effect of model sizes.} 
We present a comparison between TIM and instruction tuning across LLMs with different model sizes including BLOOM-1b7, BLOOM-3b, BLOOMZ-7b-mt, LLaMA-2-7b, and LLaMA-2-13b.}
\label{fig_modeling_scaling}
\end{figure*}

\begin{figure*}[!t]
\centering
\includegraphics[width=0.9\linewidth]{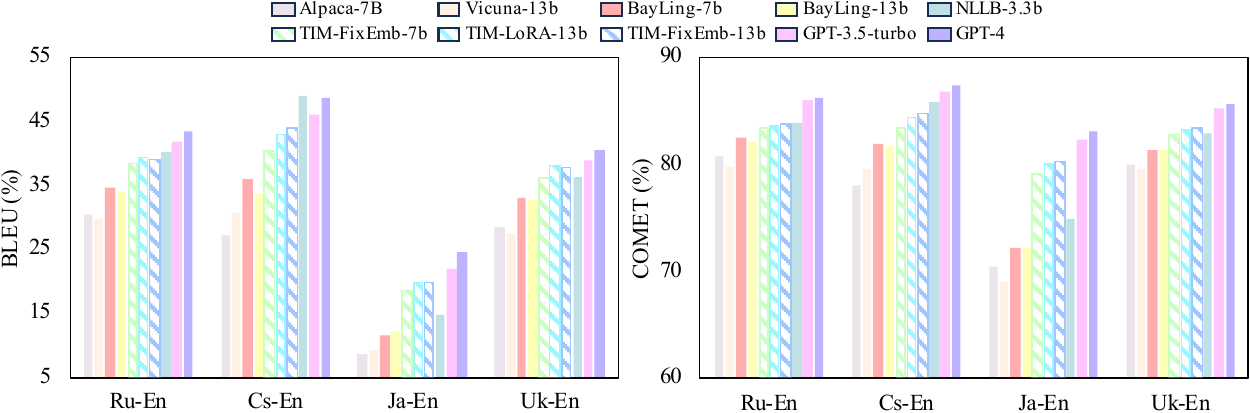}
\caption{
{Zero-shot translation.} We fine-tune LLaMA2 and compare our TIM-FixEmb-7b, TIM-LoRA-13b, and TIM-FixEmb-13b with the open-sourced models on WMT22 multilingual-to-English translation benchmark.}
\label{fig_zero_shot}
\end{figure*}

\subsection{Effect of Model Sizes}
We present a comparison between TIM and instruction tuning across different model sizes on the WMT22 test set. 

Figure \ref{fig_modeling_scaling} illustrates the consistent improvements achieved by TIM, indicating its generalizability. 
Besides, as the foundation LLM's size increases, the translation performance of the LLMs fine-tuned with TIM improves. 
In particular, the improvement is more significant when the model size is smaller. 
This observation supports our hypothesis that the smaller model has a weaker ability to comprehend instructions, and it may not effectively learn task patterns with simple instruction tuning especially using a small amount of training data, 
By contrast, training LLMs with comparison help them to better identify the task's requirements and better leverage internal cross-lingual knowledge.

\subsection{Zero-shot Translation}

To evaluate TIM’s performance in translation directions never seen previously, i.e., zero-shot multilingual capability, we conduct experiments on the WMT22 multilingual-to-English translation benchmark which encompasses 4 translation directions: 
Czech-to-English (cs$\Rightarrow$en), Japanese-to-English (ja$\Rightarrow$en), Russian-to-English (ru$\Rightarrow$en), and Ukrainian-to-English (uk$\Rightarrow$en).
We compare our method with the following open-sourced models:
Alpaca-7b\footnote{https://huggingface.co/tatsu-lab/alpaca-7b-wdiff}, Vicuna-13b\footnote{https://huggingface.co/lmsys/vicuna-13b-delta-v1.1}, BayLing-7b, -13b \cite{Arxiv2023:bayling}, NLLB-3.3b \cite{arxiv2022:NLLB}, ChatGPT, and GPT4 \cite{arxiv2023:openai_gpt4}.
We report the results of the above models in \citet{Arxiv2023:bayling}.
Due to the better performance of LLaMA-2 in multilingual-to-English, we report the performance of fine-tuned LLaMA-2-7b and LLaMA-2-13b with our TIM, respectively.

As depicted in Figure \ref{fig_zero_shot}, {\it TIM}-(*) (i.e., TIM-FixEmb-7b, TIM-LoRA-13b, and TIM-FixEmb-13b) exhibit good zero-shot multilingual capability on these translation directions. 
Compared to {\it Alpaca-7b}, {\it Vicuna-13B}, {\it BayLing-7b}, and {\it BayLing-13b}, 
{\it TIM}-(*) exhibits superior translation ability, highlighting that aligning training languages strengthens the alignment of other languages as a by-product. 
Additionally, 
{\it TIM}-(*) obtains comparative performance with {\it NLLB-3.3B} in most language pairs, and significantly better on Ja$\Rightarrow$En.
These results demonstrate that adding carefully constructed translation data, combined with an effective training strategy such as our proposed TIM, can enhance the overall task capability of LLMs.

\subsection{Ablation Study}

To analyze the impact of different components of TIM, we investigate variants of {\it TIM-FixEmb} taking {\bf BLOOMZ-7b-mt} as the backbone: 
{\it MT w/ {\rm (*)}}, where we add the (*)-guided comparisons in training data; 
{\it TIM[*]}, where we use {\it noisy-based} or {\it LM-based} bad output for preference comparison; 
{\it TIM w/o $L_{pl}$}, where we remove $\mathcal{L}_{pl}$; 
and {\it TIM w/o OutCom}, where we remove output comparison.
As a supplement to BLEU, we analyze the phenomenon of hallucination on the Zh$\Rightarrow$En test set using the hallucination detector provided by \citet{zhou-etal-2021-detecting}. 
The BLEU scores, sentence-level, and token-level hallucination scores are reported in Table \ref{tab_results_ablation}.

\begin{table}[!t]
\centering
\small
\setlength{\tabcolsep}{1.4mm}{
\begin{tabular}{l|lcccc}
\toprule
{\bf Id} & {\bf Method}
& {\bf BLEU$\uparrow$} & {\bf S-Hal.$\downarrow$} & {\bf T-Hal.$\downarrow$} & {\bf $\Delta\%$ T-Hal.} \\
\midrule
0 & Alpaca & 10.96 & 73.87 & 20.36 & - \\
1 & MT & 23.08 & 68.21 & 10.58 & -9.78\% \\
2 & \ \ {\it w/ Rev} & 23.41 & 67.36 & 9.62 & -10.74\% \\
3 & \ \ {\it w/ Dict} & 23.73 & 66.77 & {8.93} & -11.43\% \\
4 & \ \ {\it w/ Error} & 23.94 & {66.61} & 9.59 & -10.77\% \\
5 & TIM[Noisy] & 24.11 & 67.31 & 9.39 & -10.97\% \\
6 & TIM[LM] & {\bf 24.51} & {\bf 66.03} & {\bf 8.83} & {\bf -11.53\%} \\
7 & \ \ {\it w/o $L_{pl}$} & 23.76 & 68.00 & 9.53 & -10.83\% \\
8 & \ \ {\it w/o OutCom} & 23.21 & 67.46 & 9.69 & -10.67\% \\
\bottomrule
\end{tabular}}
\label{tab_results_ablation}
\caption{
{Ablation study.} 
We fine-tune BLOOMZ-7b-mt with our TIM and report BLEU and hallucination scores on Zh$\Rightarrow$En.}
\end{table}

The experimental results of 1, 2, 3, and 4 indicate a noteworthy reduction in translation hallucination when output comparison is incorporated into language models.
Particularly, the inclusion of dictionary-guided data is crucial among various data types.
This suggests that providing translation-related information and instructing the model to generate corresponding translations during training can promote the model to produce more faithful translations.
Furthermore, the results of 1 and 8 indicate that LLMs can learn better translation output through preference comparison, even without the requirement of any output comparison data. 
Finally, although the performance of {\it TIM[Noisy]} proved to be competitive with {\it TIM[LM]} in terms of BLEU and COMET scores (Table \ref{tab_results_main_result}), the results of 5 and 6 in Table \ref{tab_results_ablation} indicate that incorporating bad examples based on actual LM errors can provide more meaningful training signals compared to artificial noisy data.

\subsection{MT Metrics Evaluation}

The preference scores can reflect the quality of the model output.
To demonstrate how well they reflect quality assessment,
we use MTME\footnote{https://github.com/google-research/mt-metrics-eval} to evaluate the performance of our preference scores on standard test sets from the WMT22 Metrics Shared Tasks in De$\Rightarrow$En and En$\Rightarrow$De. 
We compare ours with some reference-free metrics: COMET-QE \cite{rei-etal-2021-cometqe}, COMETKiwi \cite{rei-etal-2022-comet22}, UniTE-src \cite{wan-etal-2022-unite}, and HWTSC-Teacher-SIM \cite{liu-etal-2022-HWTSC}); and reference-based metrics: metricx\_xxl\_MQM\_2020 \cite{freitag-etal-2022-wmt22qe}, BLEURT-20 \cite{sellam-etal-2020-bleurt}, COMET-22 \cite{rei-etal-2022-comet22}, BLEU \cite{papineni2002bleu:}, and chrF \cite{popovic-etal-chrF}. 

\begin{table}[!t]
\centering
\setlength{\tabcolsep}{1.8mm}{
\begin{tabular}{lccc}
\toprule
\multirow{2}{*}{\bf Method} & \multirow{2}{*}{\bf Acc.} & \multicolumn{2}{c}{\bf PCCs.} \\
& & {\bf De$\Rightarrow$En} & {\bf En$\Rightarrow$De} \\
\midrule
metricx\_xxl\_MQM\_2020 & {\bf 74.56} & 48.98 & {\bf 84.69} \\
BLEURT-20 & 73.68 & 45.84 & 71.89 \\
\textit{\bf TIM-LLaMA-13b$^*$} & \textit{72.81} & \textit{50.37} & \textit{62.67} \\
COMET-22 & 72.81 & 44.63 & 77.06 \\
BERTScore & 71.05 & 43.96 & 42.82  \\
\textit{\bf TIM-BLOOMZ-7b$^*$} & \textit{69.30} & \textit{\bf 62.14} & \textit{42.59} \\
COMET-QE$^*$ & 69.30 & 44.32 & 50.21  \\
COMETKiwi$^*$ & 68.42 & 40.95 & 67.35 \\
MS-COMET-QE-22$^*$ & 68.42 & 39.49 & 53.92  \\
BLEU & 67.54 & 35.24 & 17.88  \\
chrF & 65.79 & 35.45 & 34.63 \\
UniTE-src$^*$ & 64.91 & 40.20 & 50.91  \\
HWTSC-Teacher-Sim$^*$ & 60.52 & 32.17 & 38.53  \\
\bottomrule
\end{tabular}}
\label{tab_results_mtme}
\caption{
Pearson correlation of all metrics with system-level MQM scores for De$\Leftrightarrow$En. 
Rows are sorted by the system-level pairwise accuracy across the two language pairs. 
The best results are indicated in bold.
Reference-free metrics are indicated using an asterisk.
}
\end{table}

For each pair consisting of a source sentence and the corresponding hypothesis, we wrap them with our {\bf Training Prompt}, 
and use the score of the last token in the hypothesis as the final score.
Table \ref{tab_results_mtme} shows the system-level accuracy (\textbf{Acc}) and Pearson correlations (\textbf{PCCs}). 
In particular, our {\it TIM-LLaMA-13b} and {\it TIM-BLOOMZ-7b} outperform all the reference-free metrics and achieve better Pearson correlation on De$\Rightarrow$En than others.
This demonstrates that the LLMs are implicitly a reward model that can be jointly optimized during instruction tuning \cite{arxiv2023:direct}.

\section{Related Work}
Research on machine translation based on Large Language Models (LLMs) can be divided into two categories: LLMs as interface and instruction tuning. 

The studies of using LLMs as an interface focus on empirical analysis.
For example, \citet{Hendy} evaluate ChatGPT, GPT3.5 (text-davinci-003), and text-davinci-002 in eighteen different translation directions involving high
and low resource languages.
\citet{Zhu} further evaluate four popular LLMs (XGLM, BLOOMZ, OPT and ChatGPT) on 202 directions and 102 languages, and compare them with strong supervised baselines, which provides a more comprehensive benchmark result.
Many efforts are also put into investigating translation exemplars selection strategy of in-context learning \cite{emnlp/LinMAWCSOGBDPSK22,Agrawal}. Another line of work introduces knowledge, such as word alignments extracted from a dictionary, to LLMs for better translation \cite{arxiv2023:chainofdict}.

Tuning smaller LLMs (e.g., 7B) for translation tasks is a promising direction since they are better at English than supervised translation models.
However, even for directions from other languages to English, the gap between language models fine-tuned with translation data and supervised systems is still evident \cite{Jiao_ParroT,Arxiv2023:bayling}.
Different from them, we introduce output comparison and preference comparison data and present a preference regularization to alleviate hallucination and help LLMs learn translation better.

\section{Conclusion}

We propose TIM, a training method that fine-tunes open-source large language models for the translation task with the comparison of translations.
Experiments and analyses validate the effectiveness of TIM in terms of translation quality and zero-shot translation ability.
For the reference-free MT metrics evaluation, {\it TIM-LLaMA-13b} even outperforms representative metrics like COMET and BLEURT in De$\Rightarrow$En, showing that our method can well learn the translation and evaluation jointly.
Future work can explore the use of more diverse references for output comparison, and more advanced preference learning objectives.

\bibliography{aaai24,CameraReady/LaTeX/anthology,CameraReady/LaTeX/custom}

\end{document}